%% file: main.tex
\documentclass{article}
\usepackage{spconf,amsmath,graphicx}
\usepackage{boldline}
\usepackage{subfig}

\def\L{{\cal L}}

\newcommand\blfootnote[1]{%
  \begingroup
  \renewcommand\thefootnote{}\footnote{#1}%
  \addtocounter{footnote}{-1}%
  \endgroup
}

\title{Optimizing the Consumption of Spiking Neural Networks with Activity Regularization}
\name{Simon Narduzzi$^{\star \dagger}$ \qquad Siavash A. Bigdeli$^{\star}$ \qquad Shih-Chii Liu$^{\dagger}$ \qquad L. Andrea Dunbar$^{\star}$}
\address{\small{$^{\star}$CSEM, Neuchâtel, Switzerland} \qquad 
\small{$^{\dagger}$Institute of Neuroinformatics, University of Zurich and ETH Zurich, Switzerland}}
      
\begin{document}
%
\maketitle
\begin{abstract}
Reducing energy consumption is a critical point for neural network models running on edge devices. In this regard, reducing the number of multiply-accumulate (MAC) operations of Deep Neural Networks (DNNs) running on edge hardware accelerators will reduce the energy consumption during inference.
Spiking Neural Networks (SNNs) are an example of bio-inspired techniques that can further save energy by using binary activations, and avoid consuming energy when not spiking. The networks can be configured for equivalent accuracy on a task through DNN-to-SNN conversion frameworks but their conversion is based on rate coding therefore the synaptic operations can be high.
In this work, we look into different techniques to enforce sparsity on the neural network activation maps and compare the effect of different training regularizers on the efficiency of the optimized DNNs and SNNs.

\end{abstract}

\section{INTRODUCTION}
\label{sec:intro}
\input{Introduction}

\section{RELATED WORK}
\label{sec:related}
\input{RelatedWork}

\section{METHODS}
\label{sec:methods}
\input{Methods}

\section{RESULTS}
\label{sec:results}
\input{Results}

\section{CONCLUSION}
\label{sec:discussion}
\input{Conclusion}




\vfill\pagebreak

\bibliographystyle{IEEEbib}
\bibliography{references.bib}

\end{document}

%% file: Introduction.tex
Today, deep learning models are regularly deployed at the edge, allowing local real-time decision-making, efficient preprocessing, and privacy-preserving applications. Optimizations have been developed in the past few years to allow the deployment of these networks within restricted resource environments; quantization~\cite{liang2021pruning}, pruning~\cite{hoefler2021sparsity}, distillation~\cite{gou2021knowledge}, are some of them, which are applied either during training or post-training of the neural network. Great emphasis is also put on the development of efficient accelerators, that reach competitive performance compared to CPUs and GPUs. Recent hardware accelerators include optimization techniques such as computational reduction by zero-skipping~\cite{han2016eie, parashar2017scnn}. These solutions have been developed to skip zero weight computation in convolution layers~\cite{parashar2017scnn}, fully-connected layers~\cite{han2016eie} and activations~\cite{kim2017zena}. Therefore, to make the most out of this hardware, it is necessary to train and deploy sparse neural networks.

Biological neurons use discrete spikes to compute and transmit information using spike timing and spike rates to encode information. SNNs, inspired by biological neurons, imitate this behavior and are thus closer to biological systems than conventional deep neural networks.
Additionally, the sparse nature of spikes makes SNNs more suitable for low-power inference. The interest of such systems is attested by the development of neuromorphic hardware supporting SNNs, which is an active area of academic and industrial research: Intel Loihi, Synsense DynapCNN, and IBM TrueNorth are some of the recent chips developments that consume only $1/10'000$th of the energy of traditional microprocessors.

While SNNs show attractive power trade-offs for machine learning algorithms running on the edge, methods to train them efficiently are still behind conventional techniques: their binary nature makes them untrainable using backpropagation (BP) algorithms. Variations of the BP algorithm have been recently developed, allowing precise spike-timing learning, but the accuracy of DNNs is still not obtained~\cite{neftci2019surrogate, mirsadeghi2021stidi}. Therefore, conversion techniques have been developed to allow the transformation of BP-trained DNNs to SNNs~\cite{rueckauer2017conversion, sorbaro2020optimizing}, bringing SNNs on par with state-of-the-art deep neural networks. 

As a key property of SNNs, recent techniques have looked into incorporating sparsity inside the pre-conversion training for SNNs.
This is mainly enforced by an activation regularization during the DNN training.
It has been shown empirically that this regularization leads to higher sparsity of spikes after the DNN-to-SNN conversion~\cite{sorbaro2020optimizing}. Due to complex optimizations in the conversion step, the exact effect of the regularizer still remains unknown in the final implementation of the SNN.
Moreover, the effect of the regularization function itself has not been investigated in the context of stochastic optimization of the neural networks.
In this work, we investigate different regularization techniques for sparsity and their effects on training DNNs. Additionally, we look into their influence in sparsity for SNNs after the conversion.

The rest of the paper is organized as follows:
Section \ref{sec:related} presents a brief overview of prior work in sparsity training methods for neural networks.
Section \ref{sec:methods} describes the regularization techniques and metrics used in our setup.
Section \ref{sec:results} introduces our experimental scheme and results.
Section \ref{sec:discussion} summarizes the work of this paper and describes future research directions.

%% file: RelatedWork.tex
Efforts have been made toward the sparsification of deep neural networks to reduce the memory footprint of the models deployed at the edge. Weight sparsification is achieved mainly through pruning methods~\cite{louizos2017learning}. Pruning entire feature maps have also been studied~\cite{liu2017learning, kurtz2020inducing, georgiadis2019accelerating} to remove redundant information and subsequently reduce network computes. In SNNs, spikes and synaptic computation reduction are mostly exploited through temporal and spatial sparsity. Temporal sparsity of SNNs have inspired training techniques in deep learning \cite{yousefzadeh2021training, 8890681}, targeting time-series applications. 

Recently, regularization techniques have been applied to SNN training \cite{zhao2021spiking, pellegrini2021low} to increase spatial sparsity, but these do not rely on the regularization of BP-trained DNNs prior to SNN conversion. Sorbaro et al.\cite{sorbaro2020optimizing} proposes a loss to optimize the number of synaptic operations (SynOps), applied to a DNN, acting as $L_1$ regularization on activation and weights of each neuron. Spike count reduction using $L_2$ regularization on activity maps have been studied by \cite{neil2016learning} to reduce the number of spikes of the converted models. However, true sparsity should be obtained using $L_0$ regularization, which expresses the exact number of zeros in the neural network activation map. In our work, we look at various sparsity objectives (Section~\ref{sec:regularizers}), including surrogate and Hoyer~\cite{yang2019deephoyer} approximations and compare their performance on two DNNs and SNNs.

%% file: Methods.tex

\subsection{Metrics}

The efficacy of a DNN is expressed as Effective FLOPS (EFLOPS). Assuming that smart and efficient hardware platforms exist and perform computation based only on non-zero activations and non-zero weights, the EFLOPS metric describes the exact number of MAC operations performed. Therefore, the energy savings of running the network on this system instead of on a classical accelerator will be proportional to the level of sparsity in the network.
Biases are included in the EFLOPS if they are non-zero. We can then express the number of effective FLOPS used to compute an activation of the layer ${\ell}$ of a network as:
\begin{equation}
    EFLOPS_{\ell} =  \phi(W_{\ell}) \times \phi(A_{{\ell}-1}) + \phi(B_{{\ell}}),
\end{equation}
where $W_{\ell}$ is the weight matrix, $A_{\ell-1}$ is the input activation map, $B_{\ell}$ is the bias, and function $\phi(x) := x \neq 0 $ outputs a mask with ones, where its input has non-zero values.

\subsection{Regularizers}
\label{sec:regularizers}

\begin{figure*}[t!]
     \centering
     \includegraphics[width=\linewidth]{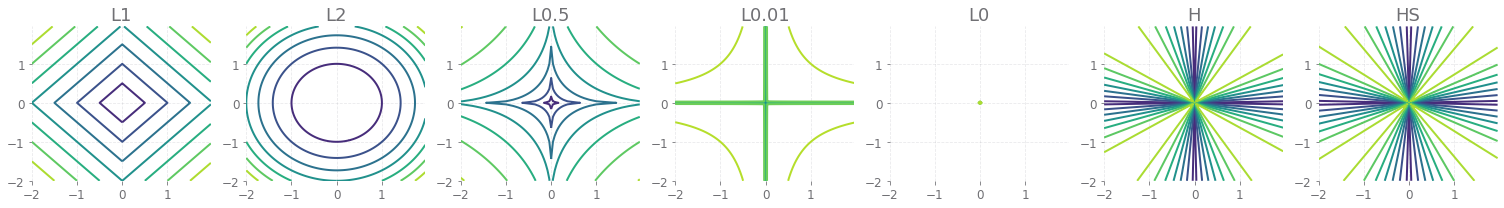}
     \caption{Loss landscape for regularization methods used in our experiment.}
     \label{fig:reg_surfaces}
\end{figure*}

We measure the activity of the network of each layer, which correlates with the SynOps emitted by the network when converted to a spiking version \cite{sorbaro2020optimizing, neil2016learning}. We denote the activation value of neuron $i$ in a layer as $x_i$. Several regularizers are tested to evaluate their capacity to generate sparse activation maps. The regularization is applied using normalization of the activation. We selected a few regularizers, whose 2D landscapes are displayed in Fig.~\ref{fig:reg_surfaces}:
\vspace{-5pt}
\paragraph*{\boldmath $L_2$}: Also known as the Euclidian norm, has already demonstrated its effectiveness in the reduction of the number of spikes \cite{neil2016learning}. While the gradients of this norm are good at high values, they tend to be reduced as the values approach~zero.
\vspace{-5pt}
\paragraph*{\boldmath $L_1$}: Also known as the Manhattan Distance, $L_1$ norm is the sum of all neurons values in the output activation map. Gradients of $L_1$ are the same everywhere and tend to have better sparsifying power than $L_2$ as the penalty for high and low values is the same.
\vspace{-5pt}
\paragraph*{\boldmath $L_0$}: Indicates the number of non-zero elements in a vector. Therefore, reducing this value of the norm yield to more zero in the activation map. We hypothesize that this norm has the best sparsifying power. However, it has non-informative gradients, making it unusable during the optimization.
\vspace{-5pt}
\paragraph*{\boldmath $L_{p<1}$}: We use a surrogate $L_0$ in the form of $L_p$, where $p$ is a positive number smaller than one. This regularization has the advantage of being differentiable everywhere, and the speed of approach (gradient) towards 0-values can be controlled using the value $p$. $L_p$ regularization is defined as:
\vspace{-5pt}
\begin{equation}
    L_{p}(X) = \sum_i{|x_i|^p}.
\end{equation}
\vspace{-25pt}
\paragraph*{Hoyer ($H$) \textmd{and} Hoyer-Squares ($H_S$)}: The Hoyer regularization is the ratio between $L_1$ and $L_2$ norm. Hoyer-Squares normalization is an approximation of the $L_0$ norm which has the advantage of being scale invariant as for the $L_0$ norm:
\begin{equation}
    H(X) = \frac{\sum_i |x_i|}{\sqrt{\sum_i x_i^2}},
\quad \text{and} \quad 
    H_S(X) = \frac{(\sum_i |x_i|)^2}{\sum_i x_i^2}.
\end{equation}
Finally, we define our loss function as:
\begin{equation}
    \L= CE + \lambda_{reg} \sum_{{\ell}}{\psi(X_{\ell})},
\end{equation}
where $CE$ is the cross-entropy loss, $\lambda_{reg}$ controls the regularization weight, $X_{\ell}$ is the activation map of layer ${\ell}$ in the neural network, and $\psi \in \{L_1, L_2, L_p, H, H_S\}$ indicates the chosen regularization function. In this work, we evaluate the reduction in number of computes relative to the value $\lambda_{reg}$.







\subsection{Network architecture and training procedure}
To assess the properties of regularization algorithms on the computational complexity of neural networks, two models were trained: LeNet-5, a convolutional neural network, and a $784$-$300$-$100$-$10$ multi-layer perceptron (MLP), both with ReLU activations in hidden layers. The models were trained with different regularizers on the MNIST dataset with Adam optimizer, a learning rate of $1e-4$, and batch-size of $128$. $5000$ images from the training set are kept for validation. The convergence criterion is set to be the smallest validation loss, with the patience of $20$ epochs.
The model is then converted to a SNN using the procedure described in \cite{rueckauer2017conversion}. The SNN model is calibrated on the full training set and the accuracy is reported on the entire test set. The converted model is simulated for $100$ time-steps, and the test accuracy at the end of the simulation is reported as the SNN accuracy. The reported EFLOPS, number of spikes and SynOps are averaged per each sample over the course of the simulation. We repeated the experiments with several random seeds and obtained similar results. Depending on the hardware, the consumption might be impacted by both spike emission and SynOps. Therefore, we report these two values independently. SynOps are however considered to be the dominant source of the power. We demonstrate that our approach can be applied to other datasets by training and testing LeNet-5 on the CIFAR-10 dataset with the same procedure, and exploring a few $\lambda_{reg}$ values. In order to get results comparative to DNNs, we increase the simulation time-steps to $1000$ in this experiment.

%% file: Results.tex

\begin{figure*}[t!]
     \centering
     \includegraphics[width=1.0\linewidth]{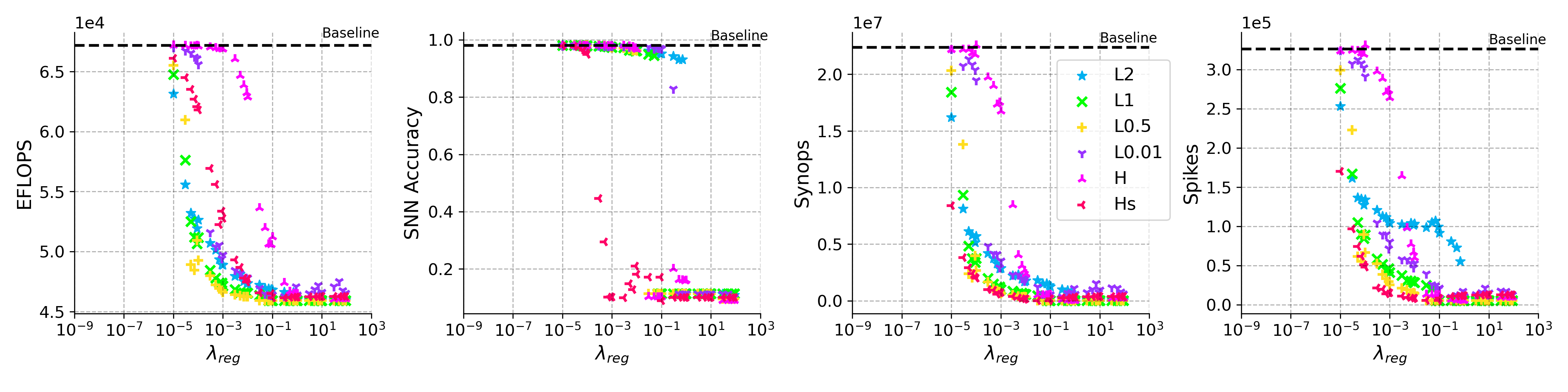}
     \caption{Metrics relative to the regularization constant $\lambda_{reg}$ in the converted SNN MLP architecture.}
     \label{fig:cross_metrics}
\end{figure*}

\subsection{Activity regularization effect on number of SynOps}
\label{subsec:mlp}

In Fig.~\ref{fig:cross_metrics}, we show the results of the regularization on the different metrics using the MLP trained on MNIST. The regularizers have different dynamics, but we observe a general trend toward the reduction of activity with increasing values of the regularization constraint, $\lambda_{reg}$. When $\lambda_{reg}$ is too high, we observe a significant accuracy drop, as the network can not learn anymore. The results of the best trade-off models are represented in Table~\ref{tab:results}. For each regularization method, we take the best model with less than $0.5\%$ and $5\%$ difference in accuracy after conversion for MNIST and CIFAR-10 respectively, with the least number of SynOps. We observe that activity regularization greatly reduces the number of spikes and SynOps of the converted network, irrespective of the chosen regularizer.
On MNIST, the total number of spikes is reduced by more than $90\%$ and the number of SynOps by $96\%$ on the MLP using $L_{0.5}$. For the LeNet-5 architecture, $L_1$ achieves the best reduction in both SynOps and number of spikes. $L_2$ still reduces the number of SynOps by $90\%$ while keeping the accuracy of the converted SNN the same. We observe that $L_{0.01}$ might be to aggressive for the models, and is worse than $L_1$ in both models. The reduction of SynOps in CIFAR-10 experiment is less significant than in MNIST. This can be due to the input (natural images) being less sparse than MNIST samples. We did not test Hoyer regularizers on CIFAR-10 because of its poor performance on MNIST.
\begin{table}[b!]
\centering
\resizebox{1.0\linewidth}{!}{\begin{tabular}{lcccc}
\hlineB{3}
& \textbf{Spikes}  & \textbf{EFLOPS} & \textbf{SynOps} & \textbf{Accuracy} \\ 
& & & & (DNN/SNN) \\
\hlineB{3}
MLP baseline & 326'185 & 67'183 & 22'374'991 & 97.91\% / 97.98\% \\
\hline
\textbf{Reg. @ $\lambda_{reg}$}  & \textbf{$\%$Spikes}  & \textbf{$\%$EFLOPS} & \textbf{$\%$SynOps} & \textbf{$\Delta$Accuracy} \\
\hline
L2 @ 9e-04 & -67.29\% & -27.20\% & -87.29\% & -0.29\%/-0.40\% \\
L1 @ 7e-04 & -86.65\% & -29.46\% & -94.61\% & -0.34\%/-0.44\% \\
L0.5 @ 9e-04 & \textbf{-90.93\%} & \textbf{-30.42\%} & \textbf{-96.37\%} & -0.30\%/-0.43\% \\
L0.01 @ 7e-03 & -83.83\% & -28.70\% & -91.15\% & -0.38\%/-0.44\% \\
H @ 7e-03 & -76.33\% & -4.91\% & -86.14\% & 0.14\%/-0.45\% \\
Hs @ 3e-05 & -70.29\% & -3.98\% & -82.94\% & 0.10\%/-0.25\% \\
\hlineB{3}
LeNet-5 baseline & 4'600'497 & 307'599 & 258'831'814 & 98.72\% / 98.39\% \\
\hline
\textbf{Reg. @ $\lambda_{reg}$} & \textbf{$\%$Spikes}  & \textbf{$\%$EFLOPS} & \textbf{$\%$SynOps} & \textbf{$\Delta$Accuracy} \\
\hline
L2 @ 1e+00 & -83.89\% & -77.85\% & -90.14\% & 0.03\%/0.10\% \\
L1 @ 1e-02 & \textbf{-89.72\%} & -77.43\% & \textbf{-93.97\%} & 0.12\%/-0.47\% \\
L0.5 @ 3e-04 & -89.03\% & -78.94\% & -91.18\% & 0.09\%/0.18\% \\
L0.01 @ 3e-04 & -87.86\% & \textbf{-79.24\%} & -88.10\% & -0.18\%/-0.43\% \\
H @ 9e-03 & -60.32\% & -43.63\% & -60.37\% & 0.36\%/-0.18\% \\
Hs @ 1e-05 & -37.58\% & -36.76\% & -42.89\% & 0.28\%/0.44\% \\
\hlineB{3}
\end{tabular}}
\caption{Results for MLP and LeNet-5 on MNIST.} 
\label{tab:results}
\end{table}

\begin{table}[b!]
\centering
\resizebox{1.0\linewidth}{!}{\begin{tabular}{lcccc}
\hlineB{3}
& \textbf{Spikes}  & \textbf{EFLOPS} & \textbf{SynOps} & \textbf{Accuracy} \\ 
& & & & (DNN/SNN) \\
\hlineB{3}
LeNet-5 baseline & 61'231'514 & 613'508 & 3'425'593'226 & 57.92\% / 51.74\% \\
\hline
\textbf{Reg. @ $\lambda_{reg}$} & \textbf{$\%$Spikes}  & \textbf{$\%$EFLOPS} & \textbf{$\%$SynOps} & \textbf{$\Delta$Accuracy} \\
\hline
L2 @ 1e-01 & \textbf{-76.48\%} & -21.24\% & \textbf{-81.48\%} & 2.29\%/2.76\% \\
L1 @ 1e-03 & -68.98\% & -15.64\% & -72.36\% & 4.79\%/5.11\% \\
L0.5 @ 1e-05 & -15.28\% & -1.17\% & -15.22\% & 1.16\%/1.94\% \\
L0.01 @ 1e-03 & -61.72\% & \textbf{-23.14\%} & -57.65\% & -4.01\%/-3.60\% \\
\hlineB{3}
\end{tabular}}
\caption{Results for LeNet-5 on CIFAR-10.} 
\label{tab:results}
\end{table}

\begin{table*}[h!]
\centering
\resizebox{0.75\linewidth}{!}{
\begin{tabular}{lcccccc}
\hlineB{3}
\textbf{Architecture} & \textbf{L2} & \textbf{L1} & \textbf{L0.5} & \textbf{L0.01} & \textbf{H} & \textbf{Hs} \\
\hline
LeNet-5 & 0.944/0.945 & 0.966/0.966 &  \textbf{0.972}/ \textbf{0.973} & 0.854/0.937 & 0.807/0.809 & 0.844/0.844 \\
MLP & 0.965/0.965 & 0.973/0.973 &  \textbf{0.974}/ \textbf{0.975} & 0.936/0.970 & 0.922/0.928 & 0.936/0.939 \\
\hline
Average  & 0.954/0.955 & 0.970/0.970 & \textbf{0.973}/\textbf{0.974} & 0.895/0.954 & 0.865/0.869 & 0.890/0.891\\
\hlineB{3}
\end{tabular}
}
\caption{Area under the curve (AUC) for both architectures. The values are formatted as (AUC (original) / AUC (smoothed)).}
\label{tab:auc}
\end{table*}

\begin{figure*}[t!]
    \centering
      
    {{\includegraphics[width=6.8cm]{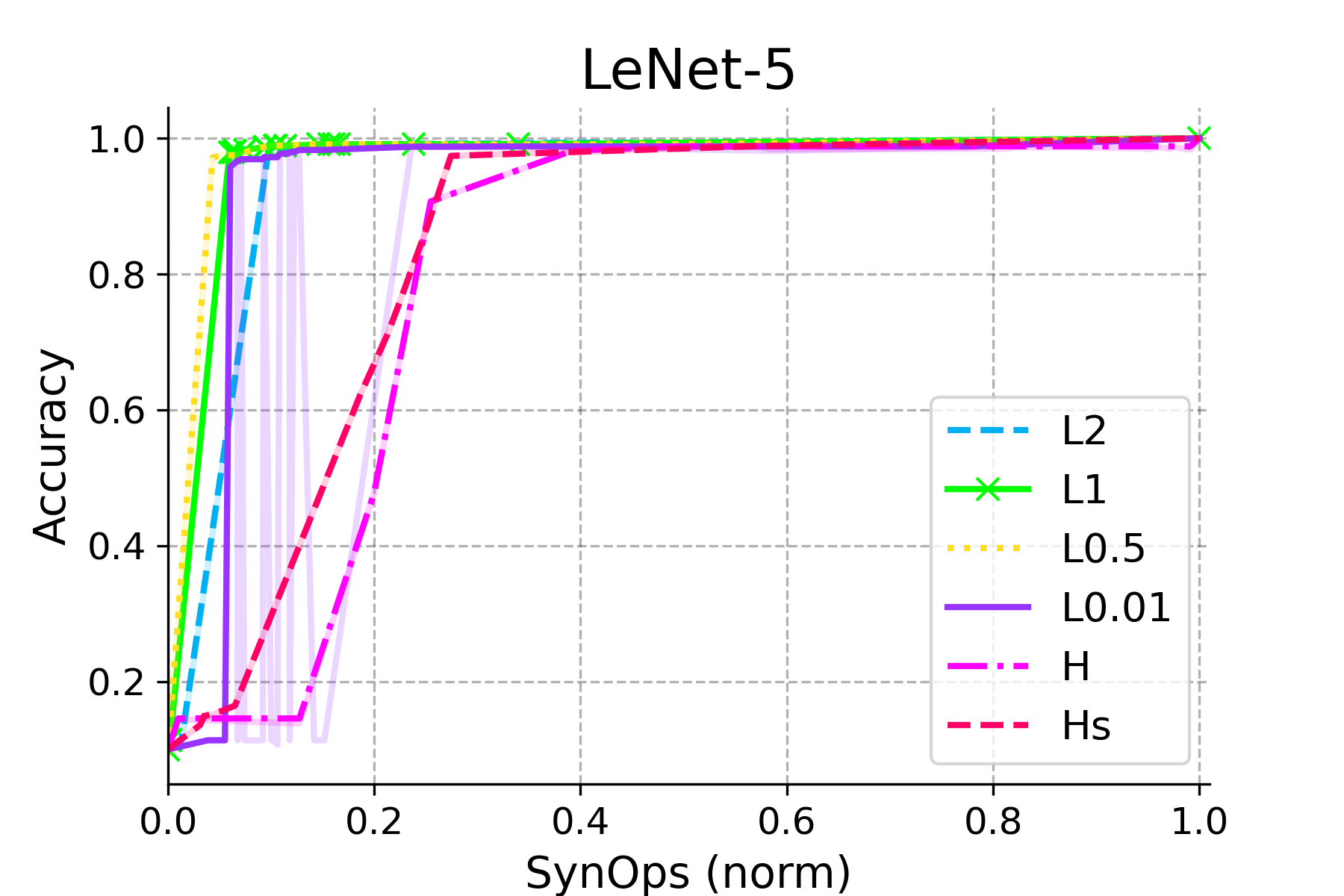} }}%
    \qquad
    {{\includegraphics[width=6.8cm]{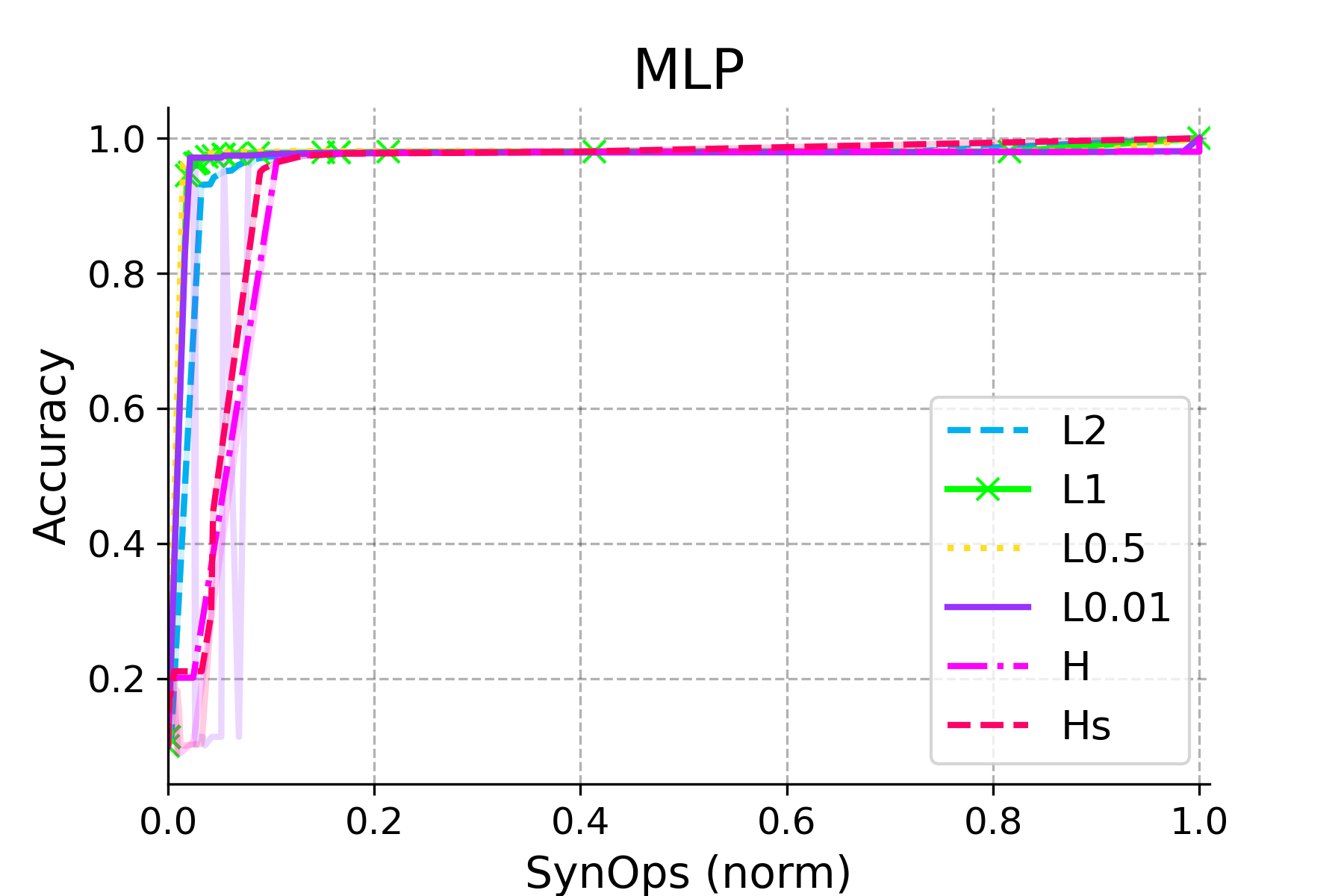} }}%
    \caption{ Accuracy vs SynOps curves for each model. SynOps are normalized to the maximum value across the models.} %
    \label{fig:curves}%
\end{figure*}

We observed that sometimes, applying regularization results in better accuracy than the baseline. However, on edge devices, an improvement of $0.2\%$ in accuracy is less desirable than an $\geq 94\%$ improvement in energy consumption due to the number of computes, as edge devices are designed for low-power and high speed rather than accuracy. While the number of SynOps is reduced, there is no correlation with the number of EFLOPS computed in the model. This suggests that SynOps reduction via pre-conversion training is more complex than only optimizing a DNN network and that artefacts are introduced during the conversion.

\subsection{$L_p$ vs. Hoyer regularization}
To further compare the performance of each metric against the other, we plot the accuracy versus the number of SynOps. As the SynOps reduction is not always monotonic (especially in the LeNet-5 architecture), the smoothed version of the curves is computed. The x-axis is normalized to the maximum number of SynOps obtained by the regularized models. Fig.~\ref{fig:curves} shows the curves for the models trained on MNIST, and Table~\ref{tab:auc} provides the corresponding Area Under the Curve (AUC) values of the original and smoothed version of the curves. We observe that Hoyer-based regularizers produce more SynOps than $L_p$ for the same accuracy in both models. $L_{p>0.01}$ have almost identical sparsifying power. We also observe the instability of the $L_{0.01}$ regularizer, where the accuracy oscillates when the SynOps become too small, suggesting that the regularization was too strong.
Regularization seems to be less effective on the LeNet-5 architecture trained on CIFAR-10, suggesting that it requires more SynOps to express the features of this dataset.

%% file: Conclusion.tex
Our results demonstrate that activity regularization during the training of DNNs is a simple way to reduce the number of spikes and SynOps in converted SNNs.
We also show that Hoyer regularization has limited effect compared to $L_p$ regularization and that reducing $p$ does not necessarily lead to better results.
This suggests that a better approximation of $L_0$ with smaller gradients than $L_{0.01}$ can give more stability and potentially even better performance.
Another potential improvement is to explore the simultaneous regularization of both weights and activations to generate very efficient SNNs: EFLOPS have not been reduced much in these models, as the weights were not constrained and could have been non-zero, potentially leading to high activations.
Finally, fine-tuning and regularization on the post-converted networks could further improve the ability of the presented methods to obtain further sparsity while keeping competitive accuracy.
\blfootnote{This project was partially funded by EU H2020, through ANDANTE \\ grant no. 876925.}